\documentclass{sigkddExp}

\usepackage{graphicx}
\usepackage{subfig}
\usepackage{amsmath}
\usepackage{footnote}
\usepackage{enumitem}
\usepackage{arydshln}
\usepackage{booktabs}
\usepackage{multicol}
\usepackage{multirow}
\usepackage{color}
\usepackage{xcolor}     
\usepackage{colortbl}
\usepackage{bbding}
\usepackage{makecell}
\usepackage{mathtools}
\usepackage{tabularray}
\usepackage{tikz}
\usepackage{url}
\usepackage{hyperref}

\definecolor{citation}{RGB}{10,110,150}  
\usepackage{hyperref}
\hypersetup{
    colorlinks=true,
    linkcolor=citation,
    urlcolor=citation,
    anchorcolor=citation,
    citecolor=citation,
}

\usepackage{tikz}
\usepackage[edges]{forest}
\definecolor{linecolor}{rgb}{0.08, 0.38, 0.74}
\definecolor{lightcoral}{rgb}{0.94, 0.5, 0.5}
\definecolor{lightgreen}{rgb}{0.56, 0.93, 0.56}
\definecolor{brightlavender}{rgb}{0.75, 0.58, 0.89}

\newcommand{\mya}{\cellcolor{gray!60!red!10!white}}
\newcommand{\myb}{\cellcolor{gray!60!blue!10!white}}
\newcommand{\myc}{\cellcolor{gray!60!green!10!white}}

\begin{document}
%

\title{\includegraphics[height=1em]{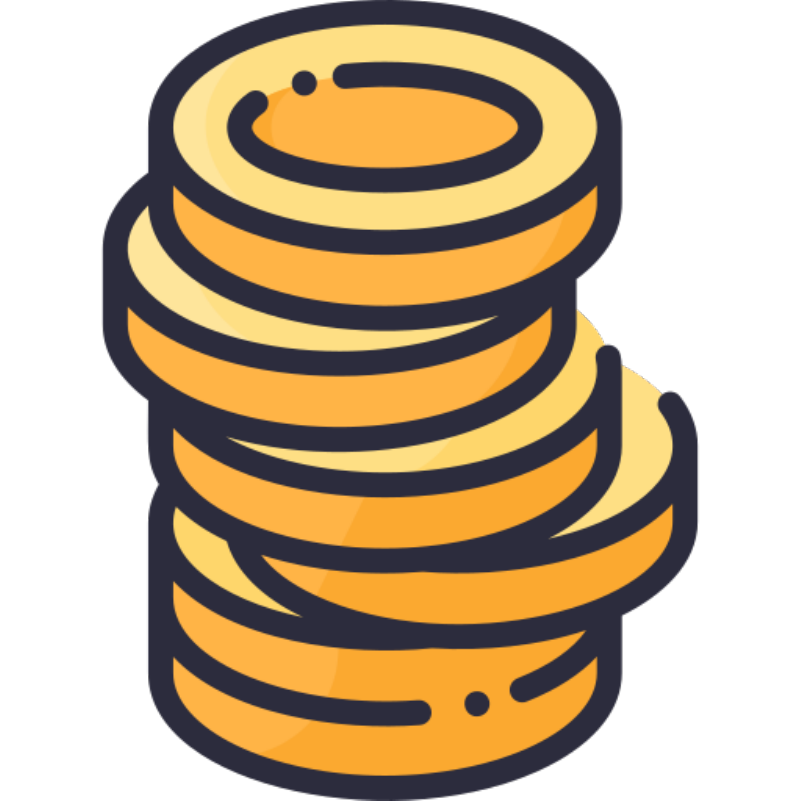} Beyond Tokens: A Survey on Decoding Methods for Large Language and Vision-Language Models}
%

\numberofauthors{3}
%


\author{Haoran Wang$^1$, Xiongxiao Xu$^2$, Philip S. Yu$^3$, Kai Shu$^1$\\
$^1$ Emory University, $^2$ Illinois Institute of Technology, $^3$ University of Illinois Chicago\\
\texttt{haoran.wang@emory.edu, xxu85@hawk.illinoistech.edu, psyu@uic.edu, kai.shu@emory.edu}
}

\maketitle

\begin{abstract}
Large language models (LLMs) and large vision-language models (LVLMs) have demonstrated impressive generative capabilities, yet ensuring their outputs align with user intent is still challenging. While most existing approaches address this issue at the training stage, inference-time approaches like decoding methods offer a more efficient and scalable solution. Decoding methods control model generation by guiding token-level selection, performing sequence-level generation, or generating tokens in parallel to accelerate the process. In this survey, we identify three emerging paradigms from recent works on decoding methods for LLMs and LVLMs, provide a systematic review of these methods, highlight ongoing challenges, and discuss potential future research directions. Our goal is to underscore the efficiency and effectiveness of decoding methods and offer a practical view of their applications. Paper lists and more resources on decoding methods for LLMs and LVLMs can be found at \url{https://github.com/wang2226/Awesome-LLM-Decoding}.
\end{abstract}

\section{Introduction}
\label{sec:introduction}
Large language models (LLMs) and large language-vision models (LVLMs) have demonstrated that scaling up both model size and training datasets can greatly improve the model's generative capabilities. However, with the rise of massive foundational models such as Llama3 405B \cite{meta2024introducing} and Megatron \cite{smith2022using}, which boasts 530 billion parameters, the focus has increasingly shifted toward developing \textit{more efficient and scalable approaches} to control generation during inference time (\S~\ref{sec:enhance}). One popular approach is prompt engineering, due to its simplicity and effectiveness. However, it is highly task-specific, requiring human expertise to craft optimal prompts, and is susceptible to prompt sensitivity issues \cite{sclar2023quantifying, lu2021fantastically}, where minor variations in prompt wording can lead to significant performance differences. Other techniques, such as ROME \cite{meng2022locating} and ITI \cite{li2024inference}, modify model internals at inference time but suffer from limited generalizability and scalability. Recently, there has been increasing interest in decoding methods, which play a critical role in controlling next-token prediction and text generation.

Decoding methods have a rich history in language modeling, ranging from greedy decoding and beam search to sampling-based approaches like top-$p$ sampling \cite{holtzman2019curious}. These methods transform the vector representations produced by the model into coherent text while controlling the quality and attributes of the generated output. Recent works have shown that adopting more advanced decoding strategies can effectively \textit{mitigate hallucination} \cite{chuang2023dola, wang2024mitigating, zhu2024ibd, wang2024valid, fang2024uncertainty}, \textit{improve safety} \cite{liu2021dexperts, zhong2024rose, xu2024safedecoding}, \textit{enhance visual grounding} \cite{deng2024seeing, leng2024mitigating}, \textit{improve reasoning} \cite{xie2023self, zhu2024deductive}, and \textit{increase robustness against noisy context} \cite{shi2023trusting, kim2024adaptive, qiu2024entropy}. Beyond improving text generation, decoding methods can also \textit{provide interpretability of the models}. For instance, \cite{wang2024chain} showed that modifying the decoding process can elicit chain-of-thought reasoning paths from LLMs. Finally, a recent line of research \cite{stern2018blockwise, xia2023speculative, wang2024speculative, leviathan2023fast, chen2023accelerating} has focused on \textit{improving the efficiency of LLM and LVLM generation} by decoding multiple tokens simultaneously to accelerate generation.

In this survey, we use decoding to denote inference-time procedures that transform model output distributions into output sequences, including token selection, search strategies, and parallel generation. While decoding is operationally part of generation in autoregressive models, we distinguish it from training-time alignment and prompt engineering, and focus specifically on inference-time control mechanisms.

\tikzstyle{my-box}=[
    rectangle,
    rounded corners,
    text opacity=1,
    minimum height=3em,
    minimum width=5em,
    inner sep=3pt,
    align=center,
    fill opacity=.5,
    line width=1.2pt,
]

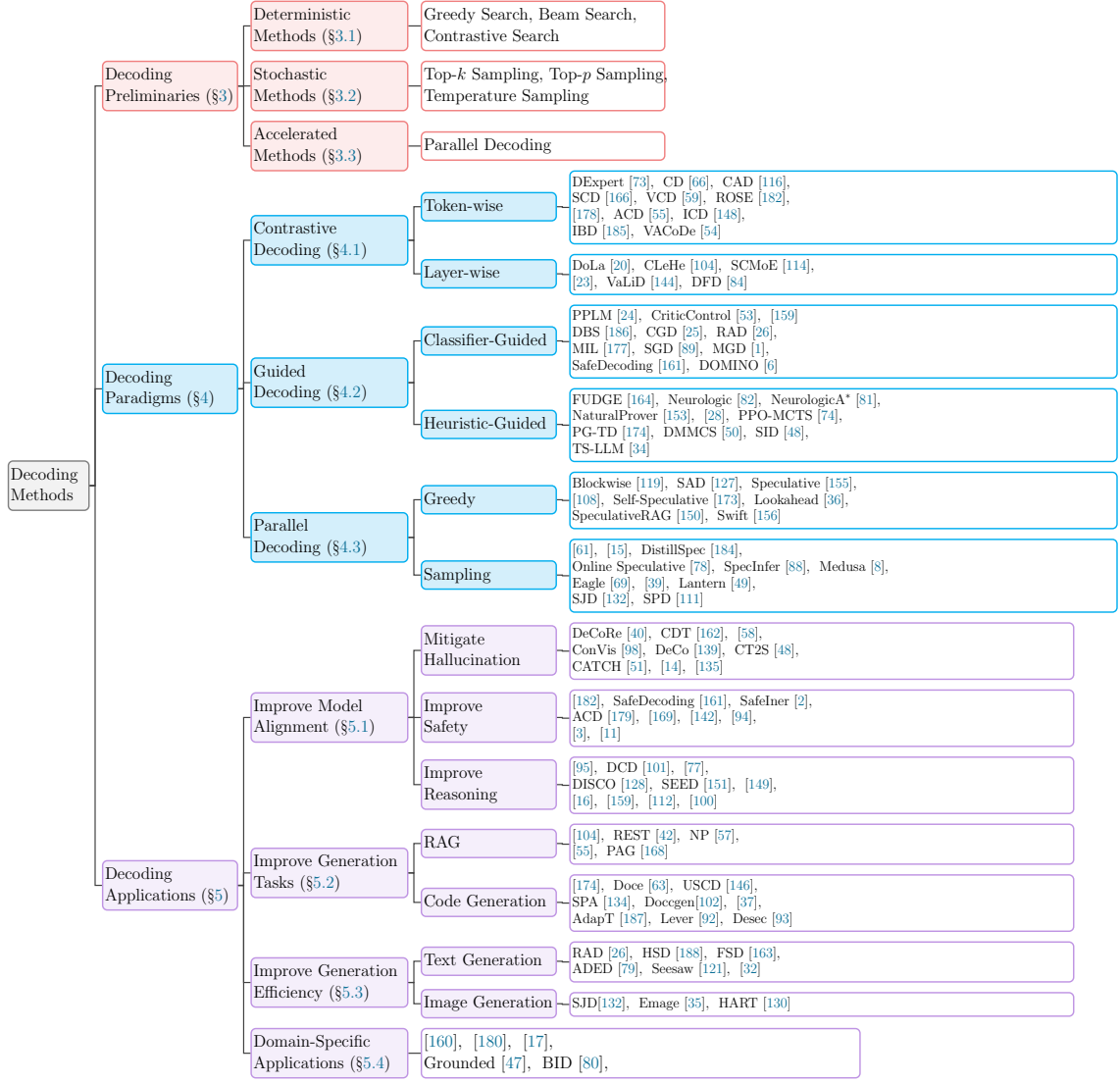
\begin{figure*}[!t]
    \centering
    \resizebox{0.85\textwidth}{!}{
        \begin{forest}
            forked edges,
            for tree={
                grow=east,
                reversed=true,
                anchor=base west,
                parent anchor=east,
                child anchor=west,
                base=center,
                font=\Large,
                rectangle,
                draw=linecolor,
                rounded corners,
                align=left,
                minimum width=4em,
                edge+={darkgray, line width=1pt},
                s sep=8pt,
                inner xsep=2pt,
                inner ysep=3pt,
                line width=1.6pt,
                ver/.style={},
            },
            where level=1{text width=12em,font=\Large}{},
            where level=2{text width=14em,font=\Large}{},
            where level=3{text width=12em,font=\Large}{},
            where level=4{text width=12em,font=\large}{},
            [
                Decoding\\Methods, draw=gray, color=gray!100, fill=gray!10, very thick, text width=7em, text=black, ver
                [
                    Decoding\\Preliminaries (\S\ref{sec:preliminaries}), color=lightcoral!100, fill=lightcoral!15, very thick, text=black, ver
                    [
                        Deterministic\\Methods (\S\ref{sec:deterministic}), color=lightcoral!100, fill=lightcoral!15, very thick, text=black
                        [
                            Greedy Search{,} Beam Search{,}\\
                            Contrastive Search
                            , color=lightcoral!100, very thick, text=black, text width=22em
                        ]
                    ]
                    [
                        Stochastic\\Methods (\S\ref{sec:stochastic}), color=lightcoral!100, fill=lightcoral!15, very thick, text=black
                        [
                           Top-$k$ Sampling{, }Top-$p$ Sampling{, }\\
                           Temperature Sampling
                            , color=lightcoral!100, very thick, text=black, text width=22em
                        ]
                    ]
                    [   
                        Accelerated\\Methods (\S\ref{sec:speculative_decoding}), color=lightcoral!100, fill=lightcoral!15, very thick, text=black
                        [
                            Parallel Decoding
                            , color=lightcoral!100, very thick, text=black, text width=22em
                        ]
                    ]
                ]
                [
                    Decoding\\Paradigms (\S\ref{sec:paradigms}), color=cyan!100, fill=cyan!15, very thick, text=black, ver
                        [
                            Contrastive\\Decoding (\S\ref{sec:contrastive}), color=cyan!100, fill=cyan!15, very thick, text=black
                            [
                                Token-wise, color=cyan!100, fill=cyan!15, very thick, text=black
                                [
                                    DExpert~\cite{liu2021dexperts}{, }
                                    CD~\cite{li2022contrastive}{, }
                                    CAD~\cite{shi2023trusting}{, }\\
                                    SCD~\cite{yuan2023speculative}{, }
                                    VCD~\cite{leng2024mitigating}{, }
                                    ROSE~\cite{zhong2024rose}{, }\\
                                    \cite{zhao2024enhancing}{, }
                                    ACD~\cite{kim2024adaptive}{, }
                                    ICD~\cite{wang2024mitigating}{, }\\
                                    IBD~\cite{zhu2024ibd}{, }
                                    VACoDe~\cite{kim2024vacode}
                                    , color=cyan!100, very thick, text=black, text width=50em
                                ]
                            ]
                            [
                                Layer-wise, color=cyan!100, fill=cyan!15, very thick, text=black
                                [
                                    DoLa~\cite{chuang2023dola}{, }
                                    CLeHe~\cite{qiu2024entropy}{, }
                                    SCMoE~\cite{shi2024unchosen}{, }\\
                                    \cite{das2024entropy}{, }
                                    VaLiD~\cite{wang2024valid}{, }
                                    DFD~\cite{luo2025odysseus}
                                , color=cyan!100, very thick, text=black, text width=50em
                                ]
                            ]
                        ]
                        [
                            Guided\\Decoding (\S\ref{sec:guided}), color=cyan!100, fill=cyan!15, very thick, text=black
                            [
                                Classifier-Guided, color=cyan!100, fill=cyan!15, very thick, text=black
                                [
                                    PPLM~\cite{dathathri2019plug}{, }
                                    CriticControl~\cite{kim2022critic}{, }
                                    \cite{xie2023self}\\
                                    DBS~\cite{zhu2024deductive}{, }
                                    CGD~\cite{deng2024seeing}{, }
                                    RAD~\cite{deng2023reward}{, }\\
                                    MIL~\cite{zhang2023mil}{, }
                                    SGD~\cite{min2024mitigating}{, }
                                    MGD~\cite{agrawal2024monitor}{, }\\
                                    SafeDecoding~\cite{xu2024safedecoding}{, }
                                    DOMINO~\cite{beurer2024guiding}
                                    , color=cyan!100, very thick, text=black, text width=50em
                                ]
                            ]
                            [
                                Heuristic-Guided, color=cyan!100, fill=cyan!15, very thick, text=black
                                [
                                    FUDGE~\cite{yang2021fudge}{, }
                                    Neurologic~\cite{lu2020neurologic}{, }
                                    NeurologicA$^*$~\cite{lu2021neurologic}{, }\\
                                    NaturalProver~\cite{welleck2022naturalprover}{, }
                                    \cite{dhuliawala2024adaptive}{, }
                                    PPO-MCTS~\cite{liu2023making}{, }\\
                                    PG-TD~\cite{zhang2023planning}{, }
                                    DMMCS~\cite{kaliosis2024data}{, }
                                    SID~\cite{huo2024self}{, }\\
                                    TS-LLM~\cite{feng2023alphazero}
                                    , color=cyan!100, very thick, text=black, text width=50em
                                ]
                            ]
                        ]
                        [   
                            Parallel\\ Decoding (\S\ref{sec:parallel}), color=cyan!100, fill=cyan!15, very thick, text=black
                            [
                                Greedy, color=cyan!100, fill=cyan!15, very thick, text=black
                                [
                                    Blockwise~\cite{stern2018blockwise}{, }
                                    SAD~\cite{sun2021instantaneous}{, }
                                    Speculative~\cite{xia2023speculative}{, }\\
                                    \cite{santilli2023accelerating}{, }
                                    Self-Speculative~\cite{zhang2023draft}{, }
                                    Lookahead~\cite{fu2024break}{, }\\
                                    SpeculativeRAG~\cite{wang2024speculative}{, }
                                    Swift~\cite{xia2024swift}
                                    , color=cyan!100, very thick, text=black, text width=50em
                                ]
                            ]
                            [
                                Sampling, color=cyan!100, fill=cyan!15, very thick, text=black
                                [
                                    \cite{leviathan2023fast}{, }
                                    \cite{chen2023accelerating}{, }
                                    DistillSpec~\cite{zhou2023distillspec}{, }\\
                                    Online Speculative~\cite{liu2023online}{, }
                                    SpecInfer~\cite{miao2023specinfer}{, }
                                    Medusa~\cite{cai2024medusa}{, }\\
                                    Eagle~\cite{li2024eagle}{, }
                                    \cite{gagrani2024speculative}{, }
                                    Lantern~\cite{jang2024lantern}{, }\\
                                    SJD~\cite{teng2024accelerating}{, }
                                    SPD~\cite{shen2024superposed}
                                    , color=cyan!100, very thick, text=black, text width=50em
                                ]
                            ]
                        ]
                ]
                [
                    Decoding\\Applications (\S\ref{sec:applications}), color=brightlavender!100, fill=brightlavender!15, very thick, text=black, ver
                        [
                            Improve Model\\Alignment (\S\ref{sec:alignment}), color=brightlavender!100, fill=brightlavender!15, very thick, text=black
                            [
                                Mitigate\\Hallucination, color=brightlavender!100, fill=brightlavender!15, very thick, text=black
                                [
                                    DeCoRe~\cite{gema2024decore}{, }
                                    CDT~\cite{yang2024improving}{, }
                                    \cite{lee2024delve}{, }\\
                                    ConVis~\cite{park2024convis}{, }
                                    DeCo~\cite{wang2024mllm}{, }
                                    CT2S~\cite{huo2024self}{, }\\
                                    CATCH~\cite{kan2024catch}{, }
                                    \cite{chen2025attention}{, }
                                    \cite{tu2025attention}
                                    , color=brightlavender!100, very thick, text=black, text width=46em
                                ]
                            ]
                            [
                                Improve\\Safety, color=brightlavender!100, fill=brightlavender!15, very thick, text=black
                                [
                                    \cite{zhong2024rose}{, }
                                    SafeDecoding~\cite{xu2024safedecoding}{, }
                                    SafeIner~\cite{banerjee2024safeinfer}{, }\\
                                    ACD~\cite{zhao2024adversarial}{, }
                                    \cite{zeng2024root}{, }
                                    \cite{wang2024probing}{, }
                                    \cite{niu2024parameter}{, }\\
                                    \cite{bao2024decoding}{, }
                                    \cite{chakraborty2024transfer}
                                    , color=brightlavender!100, very thick, text=black, text width=46em
                                ]
                            ]
                            [
                                Improve\\Reasoning, color=brightlavender!100, fill=brightlavender!15, very thick, text=black
                                [
                                    \cite{o2023contrastive}{, }
                                    DCD~\cite{phan2024distillation}{, }
                                    \cite{liu2024expediting}{, }\\
                                    DISCO~\cite{sun2024outdated}{, }
                                    SEED~\cite{wang2024seed}{, }
                                    \cite{wang2024chain}{, }\\
                                    \cite{chen2024self}{, }
                                    \cite{xie2023self}{, }
                                    \cite{shen2024learning}{, }
                                    \cite{peyrard2024era}
                                    , color=brightlavender!100, very thick, text=black, text width=46em
                                ]
                            ]
                        ]
                        [
                            Improve Generation\\Tasks (\S\ref{sec:specific}), color=brightlavender!100, fill=brightlavender!15, very thick, text=black
                            [
                                RAG, color=brightlavender!100, fill=brightlavender!15, very thick, text=black
                                [
                                    \cite{qiu2024entropy}{, }
                                    REST~\cite{he2023rest}{, }
                                    NP~\cite{lee2022nonparametric}{, }\\
                                    \cite{kim2024adaptive}{, }
                                    PAG~\cite{zeng2024planning}
                                    , color=brightlavender!100, very thick, text=black, text width=46em
                                ]
                            ]
                            [
                                Code Generation, color=brightlavender!100, fill=brightlavender!15, very thick, text=black
                                [
                                    \cite{zhang2023planning}{, }
                                    Doce~\cite{li2024doce}{, }
                                    USCD~\cite{wang2024uscd}{, }\\
                                    SPA~\cite{tian2024selective}{, }
                                    Doccgen\cite{pimparkhede2024doccgen}{, }
                                    \cite{fu2024constrained}{, }\\
                                    AdapT~\cite{zhu2023improving}{, }
                                    Lever~\cite{ni2023lever}{, }
                                    Desec~\cite{nie2024decoding}
                                    , color=brightlavender!100, very thick, text=black, text width=46em
                                ]
                            ]
                        ]
                        [
                            Improve Generation\\Efficiency (\S\ref{sec:efficiency}), color=brightlavender!100, fill=brightlavender!15, very thick, text=black
                            [
                                Text Generation, color=brightlavender!100, fill=brightlavender!15, very thick, text=black
                                [
                                    RAD~\cite{deng2023reward}{, }
                                    HSD~\cite{zhu2024hierarchical}{, }
                                    FSD~\cite{yang2023frustratingly}{, }\\
                                    ADED~\cite{liu2024adaptive}{, }
                                    Seesaw~\cite{su2025seesaw}{, }
                                    \cite{fan2025position}
                                    , color=brightlavender!100, very thick, text=black, text width=46em
                                ]
                            ]
                            [
                                Image Generation, color=brightlavender!100, fill=brightlavender!15, very thick, text=black
                                [
                                    SJD\cite{teng2024accelerating}{, }
                                    Emage~\cite{feng2023emage}{, }
                                    HART~\cite{tang2024hart}
                                    , color=brightlavender!100, very thick, text=black, text width=46em
                                ]
                            ]
                        ]
                        [   
                            Domain-Specific\\Applications (\S\ref{sec:domain}), color=brightlavender!100, fill=brightlavender!15, very thick, text=black
                            [
                                \cite{xu2024mitigating}{, }
                                \cite{zheng2024llm4brain}{, }
                                \cite{chen2024open}{, }\\
                                Grounded~\cite{huang2024grounded}{, }
                                BID~\cite{liu2024bidirectional}{, }
                                , color=brightlavender!100, very thick, text=black, text width=40em
                            ]
                    ]
                ]
            ]
        \end{forest}
    }
    \caption{Typology of decoding methods for LLMs and LVLMs.}
    \label{fig:typology}
\end{figure*}

This survey provides a comprehensive overview of advanced decoding methods for LLMs and LVLMs, highlighting their capabilities, applications, and potential. We first review inference-time generation control methods (\S~\ref{sec:enhance}) and classical decoding strategies (\S~\ref{sec:preliminaries}). We then introduce three modern decoding paradigms (\S~\ref{sec:paradigms}), followed by their applications (\S~\ref{sec:applications}). Finally, we discuss open challenges and future directions (\S~\ref{sec:challenges}). Figure~\ref{fig:typology} presents a typology of key concepts related to decoding methods in LLMs and LVLMs. Although we cover decoding methods for both LLMs and LVLMs, the current literature is substantially richer for text-based LLM decoding. Accordingly, this survey primarily emphasizes text decoding, while visual, video, and code decoding are discussed as emerging extensions.

\smallskip
\noindent \textbf{Major Paradigm Shift in Language Modeling}
As highlighted by \cite{liu2023pre}, the field of language modeling and its related tasks has undergone several paradigm shifts. Initially, there was a shift from fully supervised learning to the \textit{pre-train and fine-tune} approach \cite{radford2018improving, peters-etal-2018-deep, dong2019unified}, largely driven by the success of pre-trained LMs like BERT \cite{devlin2018bert}. More recently, there has been another major shift toward the \textit{pre-train, prompt, and predict} paradigm, where prompt engineering \cite{radford2019language, brown2020language, raffel2020exploring} has become a popular approach for adapting LMs to downstream tasks through carefully designed prompts.

\begin{figure*}[!ht]
    \centering
    \includegraphics[width=0.8\linewidth]{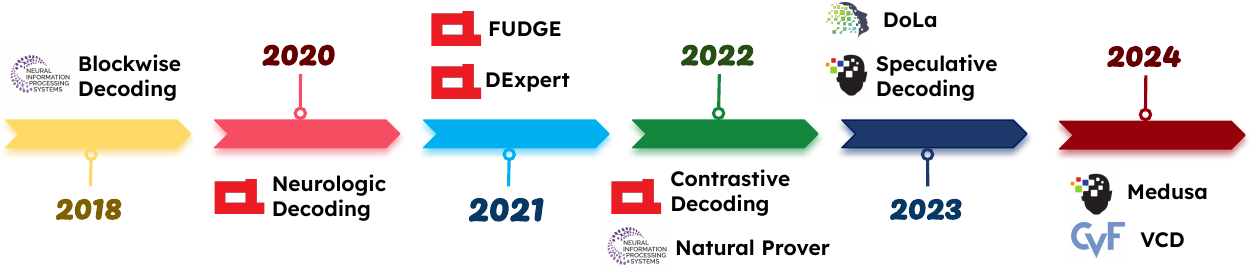}
    \caption{Timeline illustrating the evolution of decoding methods for LLMs and LVLMs.}
    \label{fig:timeline}
\end{figure*}

However, both fine-tuning and prompt engineering face challenges when applied to LLMs and LVLMs. Fine-tuning models with tens of billions of parameters requires substantial computational resources, raising concerns about energy consumption, scalability, and accessibility. While prompt engineering is computationally efficient, it remains highly task-specific and requires expert knowledge to craft effective prompts. Moreover, LLMs have been shown to have prompt sensitivity issues \cite{sclar2023quantifying, lu2021fantastically}, where even slight changes in prompt format can result in considerable variations in performance. Recently, a growing body of work has focused on advanced decoding methods \cite{dong2022survey, wiher2022decoding}. Figure~\ref{fig:timeline} illustrates the evolution of decoding methods, highlighting key milestones from classical search to modern contrastive and speculative paradigms. We argue that the field is undergoing another significant paradigm shift toward universal, effective, and scalable methods for controlling LLM generation \cite{liang2024controllable}.
\section{Inference Methods to Improve Generation}
\label{sec:enhance}
LLMs and LVLMs have demonstrated incredible performance across a wide range of NLP tasks. However, bridging the gap between their training objectives and user expectations remains challenging. While LLMs are trained to minimize contextual word prediction errors using large datasets, users expect the models to follow instructions in a helpful and safe manner. This highlights the need for alignment \cite{shen2023large, sun2024trustllm} to ensure that models behave in ways that align with human values. To achieve controllable text generation and generate aligned outputs, researchers have proposed various methods, which can be broadly categorized into training-stage and inference-stage approaches, as outlined by \cite{liang2024controllable}. Training-stage methods include fine-tuning \cite{zeldes2020technical, zhang2022discup, zhou2023controlled, zhang2023controllable, zheng2023toward, wei2021finetuned} and reinforcement learning \cite{stiennon2020learning, ouyang2022training, dai2023safe, khalifa2020distributional, upadhyay2022efficient, zeng2024token}, which leverage reward signals to guide model outputs toward specific control objectives.

In this section, we focus on methods that improve generation during the inference stage for three key reasons. First, inference-time methods enable real-time improvements without the need for re-training or altering the underlying model, making them an \textit{efficient approach} for controlling generation across models of any size. Second, these methods are typically \textit{model-agnostic}, meaning they can be applied to most decoder-only transformers, thus enhancing their versatility. Finally, some inference-time techniques like decoding methods offer better \textit{interpretability}. We categorize these inference-time approaches into three categories: \textit{(1) prompting}, \textit{(2) latent space manipulation}, and \textit{(3) decoding algorithms}.

\subsection{Prompt Engineering}
Prompt engineering directly controls text generation by crafting specific prompts for the task. The primary goal of this approach is to guide model outputs by providing clear instructions or examples. \cite{shin2020autoprompt} introduced AutoPrompt, an automated method for creating prompts across a wide range of tasks using a gradient-guided search approach. To provide a lightweight alternative to fine-tuning, \cite{li2021prefix} proposed prefix-tuning, which optimizes a sequence of continuous, task-specific vectors known as the ``prefix'' for natural language generation tasks. Additionally, \cite{lester2021power} introduced prompt tuning, a straightforward yet effective technique for learning soft prompts that condition frozen language models to perform specific downstream tasks. More recently, \cite{liu2024direct} proposed Direct Large Model Alignment (DLMA), an automatic alignment method that generates preference data using contrastive prompt pairs, calculates a self-rewarding score, and applies the DPO algorithm to align LLMs. Black-Box Prompt Optimization (BPO) \cite{cheng2023black} optimizes user prompts to align with LLMs’ input understanding, achieving the user's intent without modifying model parameters. By leveraging human preferences, BPO outperforms traditional prompt engineering in aligning LLMs with user goals. Moving beyond text, \cite{li2022align} proposed ALPRO, a video-text pre-training framework that aligns features without explicit object detectors.

\subsection{Latent Space Manipulation}
Latent space manipulation modifies the model's internal structure for controlled generation, such as attention heads or adding steering vectors to the activation layers. The core idea is that the necessary information to generate the target output is already encoded within the model’s structure, eliminating the need for re-training or fine-tuning. By operating directly on the latent space, these techniques enhance output accuracy, diversity, and coherence, while remaining computationally efficient.

\cite{chan2021deep} introduced GENhance, a generative framework that enhances attributes through a learned latent space. Additionally, \cite{subramani2022extracting} extracts latent vectors, called steering vectors, directly from PLM decoders without fine-tuning. When added to the model's hidden states, these vectors allow for controlled generation. Extending steering vectors to LLMs, \cite{turner2023activation} introduced activation engineering, a method for modifying activations during inference to steer model outputs. This approach can also be utilized to control alignment \cite{casper2024defending, wang2024inferaligner, cao2024nothing, wang2024trojan, qian2024towards}. 

To effectively control in-context learning, \cite{liu2023context} proposed a method that first creates the in-context vector from the latent embedding of the LLM, and then shifts the latent states of the LLM using these vectors to more effectively follow the demonstration examples. Furthermore, \cite{konen2024style} explores strategies to steer LLM outputs toward specific styles, such as sentiment, emotion, or writing style, by incorporating style vectors into the activations of hidden layers during text generation.

On a separate line of work, \cite{meng2022locating} analyzed the storage and recall of factual associations in autoregressive transformer language models, finding that these associations correspond to localized, directly editable computations. They modify feed-forward weights using Rank-One Model Editing (ROME) to alter factual associations within LLMs. Subsequently, \cite{meng2022mass} developed MEMIT, a method for directly updating a language model with many memories, demonstrating its ability to scale to thousands of associations for LLMs. More recently, \cite{li2024inference} introduced Inference-Time Intervention (ITI) to enhance the truthfulness of LLMs. Specifically, it shifts model activations during inference, guided by a set of directions across a limited number of attention heads.

\subsection{Decoding Algorithm}
Decoding algorithms are applied during the decoding phase of transformer-based generative models to modify the logits or probability distribution of the model’s output. It guides the generated text toward desired attributes by adjusting these probabilities, offering dynamic control over the text generation process, and ensuring the output aligns with specific requirements. Methods such as temperature scaling, top-$k$ sampling, and nucleus sampling can be used to influence the diversity, creativity, or coherence of the generated text by altering the probability distribution.

\section{Preliminaries of Decoding Strategies}
\label{sec:preliminaries}
This section reviews classical token-level decoding strategies such as greedy search, beam search, and sampling, which form the foundational building blocks for modern decoding approaches.
We define key concepts and outline the general objectives of text generation. \textbf{Auto-regressive Language Generation} operates by predicting the next token in the sequence based on the preceding tokens. Formally, considering a sequence of tokens $w=(w_1, w_2, ..., w_t)$, the probability distribution of a word sequence can be decomposed into the product of conditional next word distributions:

\begin{displaymath}
    P(w_{1:T} | W_0) = \prod_{t=1}^{T} P(w_t | w_{1:t-1}, W_0)
\end{displaymath}

with $W_0$ being the initial context word sequence. The length $T$ of the word sequence is usually determined on-the-fly and corresponds to the timestamp $t=T$ the EOS token is generated from $P(w_t|w_{1:t-1}, W_0)$.
The decoding problem is equivalent to selecting the most probable sequence given the probability distribution.

\subsection{Deterministic Methods}
\label{sec:deterministic}
Deterministic methods generate text by selecting the continuation with the highest probability determined by the LM. However, these methods often lead to model degeneration, where the output becomes unnatural, marked by repetitive and overly predictable language. As a result, the text lacks variety and fails to reflect natural human expression.

\subsubsection{Greedy Search}
Greedy search selects the word with the highest probability as its next word $\underset{w}{\operatorname{argmax}} P(w|w_{1:t-1})$ at each timestep $t$ until reaching either an end-of-sequence (EOS) token or a maximum timestep $T$.
The primary drawback of greedy search is that it may overlook high-probability words that are accessible only after a lower-probability word. As a result, greedy decoding does not formally guarantee a global optimum of the decoding objective, as its choices are only locally optimal.
Despite its simplistic approach, greedy decoding remains a widely used generation algorithm. For example, it is employed in Google’s Gemini report \cite{team2023gemini} and is commonly available in standard language model APIs.

\subsubsection{Beam Search}
Beam search reduces the risk of overlooking high-probability word sequences by maintaining a fixed number of the most likely hypotheses (or ``beams'') at each timestep, ultimately selecting the hypothesis with the highest overall probability. While beam search consistently finds output sequences with higher probabilities than greedy search, it does not guarantee the most likely output. Beam search performs well in tasks where the desired generation length is relatively predictable, such as machine translation or summarization \cite{murray2018correcting, yang2018breaking}. However, this approach is less suited for open-ended generation tasks, like dialogue or story generation, where the desired output length can vary significantly.

Beam search is prone to generating repetitive outputs. To address the lack of diversity in the generated sequences, \cite{vijayakumar2016diverse} proposed Diverse Beam Search (DBS). This algorithm divides the beam into multiple sub-groups and introduces an inner iteration at each timestep to maximize diversity between these groups.

\subsubsection{Contrastive Search}
Contrastive search \cite{su2022contrastive, su2022contrastive-1} mitigates string repetition by penalizing the selection of previously generated token sequences. This method can suppress repetitions more effectively than beam search while utilizing a comparable amount of computational resources.

\subsection{Stochastic Methods}
\label{sec:stochastic}
Human-generated text exhibits greater variance in token probabilities, reflecting a diverse range of word choices, often unexpected. In contrast, the output from deterministic methods shows minimal variance, resulting in more predictable and potentially repetitive text. To address these limitations, stochastic approaches introduce randomness during the decoding process, leading to more diverse and natural text generation. Commonly used stochastic techniques include top-$k$ sampling, top-$p$ sampling, and temperature scaling.

\subsubsection{Top-k Sampling}
In Top-$k$ sampling \cite{fan2018hierarchical, holtzman2018learning}, the $k$ most probable next words are selected, and their probability mass is redistributed to form a new distribution. Specifically, the method first identifies the top $k$ tokens with the highest probabilities for the current sequence. The probabilities of these $k$ tokens are then normalized to sum to 1, resulting in a truncated distribution. A token is then randomly sampled from this distribution and appended to the current sequence. This process is repeated iteratively until a termination condition is satisfied. GPT-2 employed this sampling strategy, which significantly contributed to its effectiveness in story generation.

While Top-$k$ sampling is both effective and significantly more efficient than beam search, it has limitations in specific scenarios, particularly in two edge cases. First, when the next-token distribution is widely spread and approaches a uniform distribution, Top-$k$ sampling may arbitrarily exclude numerous potentially interesting tokens, thereby reducing the diversity of the generated text. Conversely, when the distribution is highly concentrated, Top-$k$ sampling might either include unnecessary tokens if $k$ is too large or exclude equally probable ones if $k$ is too small.

The primary challenge with Top-$k$ sampling is how to determine an optimal $k$ value. The ideal choice of $k$ can vary based on the context and the shape of the probability distribution at each step. Using a fixed $k$ value may prove too restrictive in some scenarios while being overly permissive in others.

\subsubsection{Top-p (Nucleus) Sampling}
To address the limitations of Top-$K$ sampling, where restricting the sample pool to a fixed size $K$ can lead to gibberish outputs for sharp distributions and stifle creativity for flat distributions, \cite{holtzman2019curious} proposed Top-$P$ sampling. This method selects the smallest set of words whose cumulative probability meets or exceeds a predefined threshold $p$. Rather than sampling exclusively from the top $k$ most probable words, Top-$p$ redistributes the probability mass across this dynamic set. This adaptive approach allows the size of the word set (i.e., the number of included words) to increase or decrease based on the shape of the next-token probability distribution.

\subsubsection{Temperature Sampling}
Temperature sampling is among the most commonly used decoding strategies. Determining the optimal temperature value typically involves ad-hoc experimentation tailored to the specific application. The core idea of temperature sampling is to control the ``sharpness'' of the probability distribution by introducing a temperature parameter $t$. This parameter is applied in the softmax function after the transformer's final layer to compute token probabilities. The temperature $t$ directly influences the level of randomness in the sampling process, with higher values increasing randomness and lower values reducing it.

\subsection{Speculative Decoding}
\label{sec:speculative_decoding}
In addition to output quality, decoding strategies must also consider inference efficiency due to the ever-growing size of models. One of the main drawbacks of autoregressive decoding is that its token-by-token generation can lead to increased inference latency, scaling with both the length of the generated sequence and the model's size. To accelerate inference for LLMs, speculative decoding \cite{stern2018blockwise, leviathan2023fast, chen2023accelerating} has been introduced, allowing for the simultaneous decoding of multiple tokens per step. Specifically, in each decoding step, speculative decoding first efficiently drafts multiple tokens as speculation for future decoding steps of the target LLM, and then utilizes the LLM to verify all drafted tokens in parallel. Only those tokens that meet the LLM’s verification criteria are accepted as final outputs, ensuring generation quality \cite{xia2024unlocking}.
\section{Decoding Paradigms}
\label{sec:paradigms}
Earlier decoding methods (\S~\ref{sec:preliminaries}) primarily focused on token-level diversity and fluency. More recent work extends decoding toward sequence-level control, structured guidance, and generation efficiency. In this survey, we identify three paradigms in recent decoding methods for LLMs and LVLMs: \textbf{contrastive decoding}, \textbf{guided decoding}, and \textbf{parallel decoding}. Table~\ref{tab:paradigm_works} lists recent works categorized by their decoding paradigms. Unlike earlier token-centric decoding methods, these paradigms emerge from recent efforts that explicitly optimize sequence-level quality, controllability, or efficiency.

\begin{figure*}[!ht]
    \centering
    \includegraphics[width=0.8\linewidth]{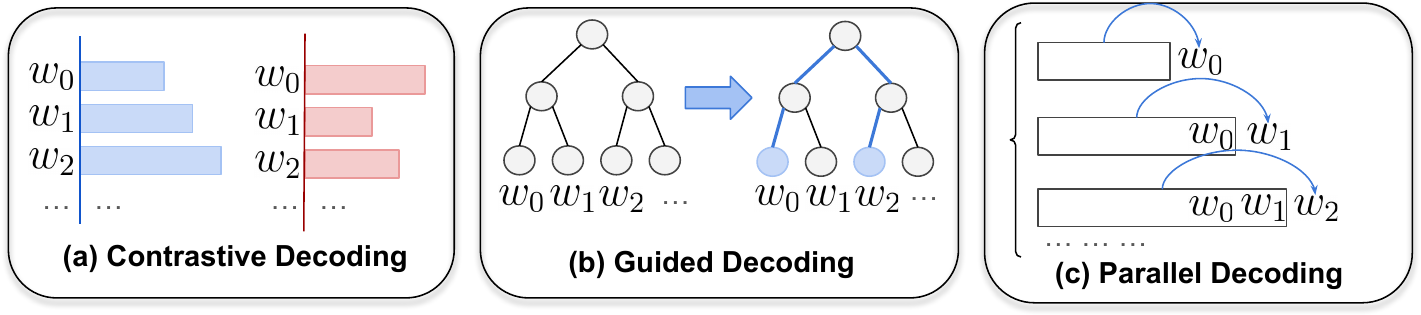}
    \caption{Illustration of different decoding paradigms. \textbf{Contrastive decoding} selects the next token by maximizing the contrast between two underlying probability distributions. \textbf{Guided decoding} determines the next token based on the highest score from a guidance function. \textbf{Parallel decoding} generates multiple token candidates simultaneously and selects the most probable one.}
    \label{fig:paradigm}
\end{figure*}

\subsection{Contrastive Decoding}
\label{sec:contrastive}
The primary goal of contrastive decoding is to enhance the quality of generated outputs by contrasting positive and negative examples during the decoding process. Unlike greedy or sampling-based decoding methods, which offer only basic control at the token level, contrastive decoding operates at both the token and layer levels, allowing for greater control over the quality of the entire generated sequence.

\textbf{Definition 1 (Contrastive Decoding).} Contrastive decoding (CD) searches for the next token that maximizes a weighted difference in likelihood between two logits $z^{+}$ and $z^{-}$. 
\begin{displaymath}
    P(W_{t} | w_{<t}) = softmax(z^{+} + \alpha (z^{+} - z^{-}))
\end{displaymath}
where $\alpha$ controls the strength of the modification.
Contrastive decoding methods can be broadly classified into two categories based on the level at which contrastive examples are utilized: \textit{token-wise CD} and \textit{layer-wise CD}.

\subsubsection{Token-wise Contrastive Decoding}
Token-wise CD contrasts the token-level probability distributions produced by pairs of contrasting examples. These contrasting examples can be model-generated. Specifically, an expert model generates logits $z^{+}$, representing the user-desired direction while a weak or base model generates logits $z^{-}$, providing a baseline probability distribution. The contrastive decoding mechanism then adjusts token probabilities by weighting the difference between $z^{+}$ and $z^{-}$, regulated by the parameter $\alpha$. For example, DExpert \cite{liu2021dexperts} utilizes both expert and anti-expert models to guide output for language detoxification and sentiment-controlled generation. During the decoding process, each token is assigned a high probability if it is deemed likely by the expert LM and unlikely by the anti-expert LM. Similarly, \cite{li2022contrastive} proposed a contrastive method that directly compares off-the-shelf LMs by computing the difference between their log probabilities.

With the increasing size of LLMs, the efficiency benefits of advanced decoding methods have become more apparent. ROSE \cite{zhong2024rose} applies contrastive decoding to pairs of carefully crafted reverse prompts to enhance LLM safety. Addressing the role of contextual information in text generation, \cite{shi2023trusting} proposed context-aware decoding (CAD), which uses a contrastive output distribution to highlight differences in token probabilities when the model generates text with and without contextual input. Building on this, \cite{zhao2024enhancing} developed a method that combines contrastive decoding with adversarial irrelevant passages as negative samples, improving robust contextual grounding by contrasting relevant and irrelevant contexts. Extending these approaches to noisy contexts, adaptive contrastive decoding (ACD) \cite{kim2024adaptive} effectively leverages contextual influences to address challenges posed by noisy or imperfect inputs. To further enhance the efficiency of contrastive decoding, \cite{yuan2023speculative} introduced Speculative Contrastive Decoding (SCD), a simple yet powerful approach that utilizes predictions from smaller LMs to achieve faster decoding while maintaining generation quality.

In the context of LVLMs, various CD methods have been proposed to enhance accuracy and mitigate hallucination. Visual Contrastive Decoding (VCD) \cite{leng2024mitigating} ensures visual grounding by comparing outputs from original and distorted visual inputs. Image-biased decoding (IBD) \cite{zhu2024ibd} contrasts predictions from conventional and image-biased LVLMs to improve image-text alignment and reduce hallucinations. Instruction Contrastive Decoding (ICD) \cite{wang2024mitigating} refines outputs by comparing distributions from standard and perturbed instructions, filtering hallucinated concepts. More recently, Visual Augmented Contrastive Decoding (VACoDe) \cite{kim2024vacode} introduced an adaptive approach, selecting optimal augmentations per task via a novel softmax distance metric, surpassing single-augmentation methods.

\subsubsection{Layer-wise Contrastive Decoding}
Recent studies have shown that transformer models tend to encode lower-level information, such as part-of-speech tags, in the earlier layers, whereas more semantic information is encoded in the later layers \cite{tenney2019bert}. For instance, \cite{dai2021knowledge} found that knowledge neurons are concentrated in the topmost layers of the BERT model. Moreover, \cite{meng2022locating} demonstrated that factual knowledge can be edited by manipulating a specific set of feedforward layers within an autoregressive model. Building on these observations, contrastive decoding has been extended to operate on layers within the models, besides just token-level adjustments. 

Decoding by Contrasting Layers (DoLa) \cite{chuang2023dola} enhances factual knowledge in LMs by leveraging modular encoding and contrastive decoding. It derives the next-token distribution by contrasting logits from different transformer layers projected onto the vocabulary space, based on the observation that factual knowledge is localized in specific layers. In a similar fashion, \cite{das2024entropy} proposed an entropy-guided method to extrapolate token probabilities beyond the last layer for more accurate contrastive decoding, instead of relying solely on the final layer. To reduce noise from retrieved context, \cite{qiu2024entropy} employed an entropy-based document-parallel ensemble decoding, which prioritizes low-entropy distributions from retrieved documents. Specifically, it compares this low-entropy ensemble with the model’s high-entropy internal distribution, emphasizing reliable external information. By leveraging the modular and hierarchical nature of factual knowledge within LLMs, Dynamic Focus Decoding (DFD) \cite{luo2025odysseus} adaptively adjusts the decoding focus based on distributional differences across layers.

More recently, \cite{shi2024unchosen} addressed the underutilization of Mixture-of-Experts (MoE) models, where unchosen experts do not contribute to the output. They proposed Self-Contrast Mixture-of-Experts (SCMoE), an inference strategy that improves performance by contrasting strong and weak expert activations within the model. For LVLMs, \cite{wang2024valid} found that key visual features in early layers often distort as they propagate to the output. Building on this finding, they introduced Visual Layer Fusion Contrastive Decoding (VaLiD), which uses uncertainty to guide visual layer selection, reducing distortions and mitigating hallucinations.

\subsection{Guided Decoding}
\label{sec:guided}
Selecting the right decoding ``path'' can greatly enhance the generation quality of LLMs and LVLMs. For example, chain-of-thought reasoning can emerge simply by modifying the decoding process \cite{wang2024chain}. To determine the optimal decoding path,'' various strategies have been proposed to guide the decoding process. We refer to these strategies collectively as guided decoding, which is defined as follows:

\textbf{Definition 2 (Guided Decoding).} Guided decoding (GD) searches for the next token that maximizes the score from a guidance function $\mathcal{G}$.
\begin{displaymath}
    P'(w_{t} | w_{<t}) \propto \mathcal{G}(P(w_{t} | w_{<t}), C)
\end{displaymath}
where $\mathcal{G}$ is a guidance function that adjusts the model's distribution based on certain criteria, and $C$ denotes the control condition.
We classify guided decoding into two main categories: \textit{classifier-guided} and \textit{heuristic-guided} decoding.

\subsubsection{Classifier-Guided Decoding}
Classifier-guided decoding utilizes an external classifier to influence the decoding process, allowing for control over specific attributes during text generation. The classifier can take various forms, such as a reward model, a pre-trained model, or an API, and adjust the model’s output accordingly.

By using an attribute model to guide the decoding process, Plug and Play Language Model (PPLM) \cite{dathathri2019plug} controls text generation by combining a pre-trained LM with one or more simple attribute classifiers. To detoxify PLMs at the token level, \cite{zhang2023mil} proposed Multiple Instance Learning Decoding (MIL-Decoding), which computes token-level toxicity scores and adjusts probabilities dynamically based on context. CriticControl \cite{kim2022critic} combines reinforcement learning and weighted decoding, using an LM-steering critic for controlled text generation. Similarly, Reward-Augmented Decoding (RAD) \cite{deng2023reward} modifies token probabilities using a unidirectional reward model, optimizing sampling for better attribute control.

To address uncertainty in multi-step reasoning, \cite{xie2023self} proposed a decoding algorithm using self-evaluation guidance through stochastic beam search, improving prediction quality by enhancing search efficiency. To reduce reasoning errors in intermediate steps, Deductive Beam Search (DBS) \cite{zhu2024deductive} integrates chain-of-thought and deductive reasoning with a step-wise beam search and a verifier to check the validity of each step, reducing error accumulation. SafeDecoding \cite{xu2024safedecoding} mitigates jailbreak risks by guiding the decoding process with a trained expert model to generate helpful and safe responses. For improved code generation in LLMs, \cite{agrawal2024monitor} introduced monitor-guided decoding (MGD), where a monitor uses static analysis to guide decoding. Lastly, \cite{beurer2024guiding} proposed DOMINO, a decoding algorithm that enforces subword-aligned constraints with minimal overhead, addressing challenges in aligning sub-word tokens and grammar terminals.

For LVLMs, CLIP-Guided Decoding (CGD) \cite{deng2024seeing} enhances visual grounding by using CLIP to guide the decoding process, with CLIP similarity to the image serving as a stronger indicator of hallucinations than token likelihoods. \cite{fang2024uncertainty} introduced dropout decoding, inspired by dropout regularization, to reduce hallucinations by quantifying and masking uncertain visual tokens during inference. Summary-Guided Decoding (SGD) \cite{min2024mitigating} addresses hallucinations by shortening token length through summarization, promoting greater visual detail while controlling image-related part-of-speech tokens to maintain text quality.

\subsubsection{Heuristic-Guided Decoding}
To provide more effective guidance within the token search space, heuristics are used to steer the decoding process with specific rules or search strategies. These heuristics could include constraints, results from lookahead search, or value functions.

Future Discriminators for Generation (FUDGE) \cite{yang2021fudge} adjusts the probability distribution during generation by predicting the attribute probability of the evolving sequence and modifying the logits to align with desired attributes. To address the challenges PLMs face in adhering to constraints during generation, \cite{lu2020neurologic} proposed Neurologic Decoding, which enforces lexical constraints during decoding. \cite{lu2021neurologic} later improved it with A*-inspired lookahead heuristics for better performance. For generating mathematical proofs, \cite{welleck2022naturalprover} developed NaturalProver, a knowledge-grounded model that generates proofs by conditioning on background theorems and definitions, optionally enforcing them through constrained decoding. Lastly, \cite{dhuliawala2024adaptive} introduced Adaptive Decoding, a neural module that predicts the optimal sampling temperature for specific tasks.

Using heuristics from lookahead search results, Planning-Guided Transformer Decoding (PG-TD) \cite{zhang2023planning} leverages a planning algorithm to conduct lookahead searches and guide the model in generating more effective programs for code generation with LLMs. To leverage value models trained as byproducts when aligning LMs with human preferences, \cite{liu2023making} proposed an effective method for applying Monte-Carlo tree search decoding on top of PPO-trained policy and value models. This approach integrates the value network from PPO, enabling it to collaborate closely with the policy network during inference-time generation. More recently, TS-LLM \cite{feng2023alphazero} leverages tree search with a learned value function to guide LLM decoding, enhancing reasoning, planning, and decision-making capabilities. Integrative Decoding \cite{cheng2024integrative} improves actuality by implicitly incorporating self-consistency within its decoding objective.

To mitigate hallucinations in LVLMs, \cite{huo2024self} introduced Self-Introspective Decoding (SID) with the Context and Text-aware Token Selection (CT2S) strategy. SID retains only the least important vision tokens after the early decoder layers, adaptively addressing hallucinations in vision-and-text associations during autoregressive decoding. This approach leverages the ability of pre-trained LVLMs to introspectively evaluate the significance of vision tokens based on prior vision, text, and generated content. For diagnostic captioning, \cite{kaliosis2024data} proposed Distance from Median Maximum Concept Similarity (DMMCS), a data-driven guided decoding method that incorporates medical image tags to generate more accurate diagnostic text.

\begin{table*}[!htbp]
\small
\centering
\renewcommand\arraystretch{1.1}
\setlength{\tabcolsep}{3pt}
\caption{List of works categorized by the three paradigms identified in this survey. Based on chronological order.}
\label{tab:paradigm_works}

\resizebox{0.97\linewidth}{!}{%
\begin{tabular}{>{\centering\arraybackslash}m{1.8cm}>{\arraybackslash}m{5.2cm}>{\arraybackslash}m{6cm}>{\arraybackslash}m{1cm}>{\centering\arraybackslash}m{1.2cm}}
\toprule[1pt]
\textbf{Paradigm} & \textbf{Work} & \textbf{Description} & \textbf{Model} & \textbf{Year ($\downarrow$)} \\ 
\hline

\multirow{16}{*}{\makecell{Contrastive\\Decoding}}
& \mya DExpert~\cite{liu2021dexperts} & \mya Detoxification with contrastive decoding & \mya PLM & \mya 2021 \\
& CD~\cite{li2022contrastive} & Formulate Contrastive Decoding & PLM & 2022 \\
& \mya CAD~\cite{shi2023trusting} & \mya Context-aware Decoding & \mya LLM & \mya 2023 \\
& SCD~\cite{yuan2023speculative} & Speculative contrastive decoding & LLM & 2023 \\
& \mya DoLa~\cite{chuang2023dola} & \mya Decoding by contrasting layers & \mya LLM & \mya 2023 \\
& VCD~\cite{leng2024mitigating} & Visual contrastive decoding & LVLM & 2024 \\
& \mya ROSE~\cite{zhong2024rose} & \mya Reverse prompt contrastive decoding & \mya LLM & \mya 2024 \\
& Zhao et al.~\cite{zhao2024enhancing} & Multi-input contrastive decoding & LLM & 2024 \\
& \mya CLeHe~\cite{qiu2024entropy} & \mya Entropy-based contrastive decoding & \mya LLM & \mya 2024 \\
& ACD~\cite{kim2024adaptive} & Adaptive contrastive decoding & LLM & 2024 \\
& \mya SCMoE~\cite{shi2024unchosen} & \mya Self-contrast Mixture-of-Experts & \mya LLM & \mya 2024 \\
& Das et al.~\cite{das2024entropy} & Entropy guided extrapolative decoding & LLM & 2024 \\
& \mya ICD~\cite{wang2024mitigating} & \mya Instruction contrastive decoding & \mya LVLM & \mya 2024 \\
& IBD~\cite{zhu2024ibd} & Image-biased decoding & LVLM & 2024 \\
& \mya VACoDe~\cite{kim2024vacode} & \mya Visual augmented contrastive decoding & \mya LVLM & \mya 2024 \\
& VaLiD~\cite{wang2024valid} & Visual Layer Fusion contrastive decoding & LVLM & 2024 \\
\hline

\multirow{23}{*}{\makecell{Guided\\Decoding}}
& \myb PPLM~\cite{dathathri2019plug} & \myb Attribute model guidance & \myb PLM & \myb 2020 \\
& Neurologic~\cite{lu2020neurologic} & Constrained decoding & PLM & 2020 \\
& \myb FUDGE~\cite{yang2021fudge} & \myb Future discriminator guidance & \myb PLM & \myb 2021 \\
& Neurologic$A^\star$~\cite{lu2021neurologic} & Lookahead heuristics guidance & PLM & 2021 \\
& \myb CriticControl~\cite{kim2022critic} & \myb Critic-guided decoding & \myb PLM & \myb 2022 \\
& NaturalProver~\cite{welleck2022naturalprover} & Stepwise constrained decoding & PLM & 2022 \\
& \myb MIL-Decoding~\cite{zhang2023mil} & \myb Multiple instance learning guidance & \myb PLM & \myb 2023 \\
& RAD~\cite{deng2023reward} & Reward augmented decoding & LLM & 2023 \\
& \myb PPO-MCTS~\cite{liu2023making} & \myb Value-guided Monte-Carlo tree search & \myb LLM & \myb 2023 \\
& PG-TD~\cite{zhang2023planning} & Planning-guided decoding & LLM & 2023 \\
& \myb Xie et al.~\cite{xie2023self} & \myb Self-evaluation guided beam search & \myb LLM & \myb 2023 \\
& DBS~\cite{zhu2024deductive} & Decoding deducible rationale for CoT & LLM & 2024 \\
& \myb DMMCS~\cite{kaliosis2024data} & \myb Data-driven guided decoding & \myb LVLM & \myb 2024 \\
& SGD~\cite{min2024mitigating} & Summary-Guided decoding & LVLM & 2024 \\
& \myb SID~\cite{huo2024self} & \myb Self-Introspective decoding & \myb LVLM & \myb 2024 \\
& TS-LLM~\cite{feng2023alphazero} & AlphaZero-like tree-search & LLM & 2024 \\
& \myb Dropout~\cite{fang2024uncertainty} & \myb Dropout decoding & \myb LVLM & \myb 2024 \\
& MGD~\cite{agrawal2024monitor} & Monitor-guided decoding & LLM & 2024 \\
& \myb SafeDecoding~\cite{xu2024safedecoding} & \myb Expert model guidance & \myb LLM & \myb 2024 \\
& DOMINO~\cite{beurer2024guiding} & Minimally-Invasive constrained decoding & LLM & 2024 \\
& \myb CGD~\cite{deng2024seeing} & \myb Clip-guided decoding & \myb LVLM & \myb 2024 \\
& DFD~\cite{luo2025odysseus} & Dynamic Focus Decoding & LLM & 2025 \\
& \myb AttnReal~\cite{tu2025attention} & \myb Attention reallocation & \myb LVLM & \myb 2025 \\
\hline

\multirow{20}{*}{\makecell{Parallel\\Decoding}}        
& \myc Blockwise~\cite{stern2018blockwise} & \myc Blockwise parallel decoding & \myc PLM & \myc 2018 \\
& SAD~\cite{sun2021instantaneous} & Shallow aggressive decoding & PLM & 2021 \\
& \myc SpecDec~\cite{xia2023speculative} & \myc Speculative decoding for seq2seq generation & \myc PLM & \myc 2023 \\
& Santilli et al.~\cite{santilli2023accelerating} & Hybrid GS-Jacobi decoding & PLM & 2023 \\
& \myc Self-speculative~\cite{zhang2023draft} & \myc Self-speculative decoding & \myc LLM & \myc 2023 \\
& Speculative~\cite{leviathan2023fast} & Speculative sampling & LLM & 2023 \\
& \myc Speculative~\cite{chen2023accelerating} & \myc Speculative sampling with distributed serving & \myc LLM & \myc 2023 \\
& DistillSpec~\cite{zhou2023distillspec} & Speculative via knowledge distillation & LLM & 2023 \\
& \myc SpecInfer~\cite{miao2023specinfer} & \myc Tree-based speculative verification & \myc LLM & \myc 2023 \\
& Online Speculative~\cite{liu2023online} & Online Speculative & LLM & 2023 \\
& \myc Speculative RAG~\cite{wang2024speculative} & \myc Speculative RAG & \myc LLM & \myc 2024 \\
& Lookahead~\cite{fu2024break} & Lookahead decoding & LLM & 2024 \\
& \myc Medusa~\cite{cai2024medusa} & \myc Multiple decoding heads & \myc LLM & \myc 2024 \\
& Eagle~\cite{li2024eagle} & Extrapolative speculative sampling & LLM & 2024 \\
& \myc Gagrani et al.~\cite{gagrani2024speculative} & \myc Multimodal speculative & \myc LVLM & \myc 2024 \\
& Lantern~\cite{jang2024lantern} & Latent neighbor token acceptance relaxation & LVLM & 2024 \\
& \myc SJD~\cite{teng2024accelerating} & \myc Speculative Jacobi decoding & \myc LVLM & \myc 2024 \\
& SPD~\cite{shen2024superposed} & Superposed decoding & LLM & 2024 \\
& \myc Swift~\cite{xia2024swift} & \myc On-the-fly Self-speculative decoding & \myc LLM & \myc 2024 \\
& Seesaw~\cite{su2025seesaw} & Dynamic Model Resharding & LLM & 2025 \\
\bottomrule[1pt]
\end{tabular}
}
\end{table*}

\subsection{Parallel Decoding}
\label{sec:parallel}
Unlike standard sequential decoding, parallel decoding executes multiple decoding processes concurrently where it first generates multiple candidate sequences simultaneously and then selects the most likely one based on a set of predefined criteria.

\textbf{Definition 3 (Parallel Decoding).} Parallel decoding first decodes multiple future tokens $y_1, y_2, ..., y_m$ simultaneously, a process often referred to as \textit{drafting}.
\begin{displaymath}
\begin{cases}
  y_1 = \arg\max P(w_t | w_0) \\    
  y_2 = \arg\max P(y_2 | y_1, w_0) \\   
  \vdots \\
  y_m = \arg\max P(y_m | y_{1:m-1}, w_0)
\end{cases}
\end{displaymath}
Then, it aggregates these tokens $y_1, y_2, ..., y_m$ in parallel using the target LLM to speed up inference -- a process known as \textit{verification}.
\begin{displaymath}
    P(w_{t} | w_{<t}) = \mathcal{M}(w | w_{\leq t}, \hat{w}_{\leq i}), i=1, ..., K+1
\end{displaymath}
where $\mathcal{M}$ is the draft model, and $\hat{w_1}$, $\hat{w_2}$, ..., $\hat{w_K}$ are the draft tokens outputted by $\mathcal{M}$.
This \textit{draft-then-verify} paradigm can be further classified into two categories: \textit{greedy} and \textit{sampling}.

\subsubsection{Greedy} 
To enhance the decoding process in deep autoregressive models, \cite{stern2018blockwise} introduced blockwise decoding, a parallel approach in which predictions are made for multiple time steps simultaneously, followed by a rollback to the longest valid prefix as determined by a scoring model. To improve online inference efficiency in transformer-based models for instantaneous grammatical error correction, Shallow Aggressive Decoding (SAD) \cite{sun2021instantaneous} uses a shallow decoder to aggressively decode as many tokens as possible in parallel. To overcome the efficiency limitations of autoregressive decoding in transformers, \cite{santilli2023accelerating} redefine standard greedy autoregressive decoding for machine translation by adopting a parallel formulation that uses Jacobi and Gauss-Seidel fixed-point iteration methods to achieve faster inference.

Notably, \cite{xia2023speculative} introduced Speculative Decoding (SpecDec), a method to accelerate autoregressive decoding through speculative execution, or draft-then-verify. It consists of two components: Spec-Drafter, an independent model for efficient token drafting, and Spec-Verification, a mechanism for validating the drafted tokens. To enhance inference efficiency in RAG, \cite{wang2024speculative} proposed SpeculativeRAG, a framework in which a larger generalist LM verifies multiple RAG drafts generated in parallel by a smaller, distilled specialist LM. Each draft, based on distinct subsets of retrieved documents, reduces token counts and offers diverse perspectives, improving comprehension, mitigating position bias, and speeding up the RAG process by limiting the generalist LM to a single verification pass.

To eliminate the need for separate draft and verifier models, self-speculative decoding \cite{zhang2023draft} uses a single LLM for both drafting and verification, avoiding additional training and memory overhead. More recently, \cite{fu2024break} introduced Lookahead Decoding, a parallel algorithm that accelerates LLM decoding without relying on auxiliary models or data stores. This approach balances per-step log (FLOPs) with the total decoding steps, making it highly parallelizable on modern accelerators and compatible with memory-efficient attention mechanisms like FlashAttention. Lastly, \cite{su2025seesaw} proposed a technique to facilitate dynamic reconfiguration of parallelization strategies across prefilling and decoding stages to improve the efficiency of distributed LLM inference.

\subsubsection{Sampling}
Similar to how stochastic decoding methods often surpass deterministic approaches, integrating speculative sampling can significantly boost the performance of parallel decoding. Speculative sampling \cite{leviathan2023fast} speeds up sampling from autoregressive models while preserving accuracy by extending speculative execution to the stochastic setting. It enhances exact decoding from large models by running them in parallel with approximation models, generating multiple tokens concurrently without altering the underlying distribution. Similarly, \cite{chen2023accelerating} used speculative sampling to generate multiple tokens per transformer call, optimizing distributed model serving. To align compact draft models with target models, DistillSpec \cite{zhou2023distillspec} applies knowledge distillation before speculative decoding, customizing the divergence function. More recently, \cite{shen2024superposed} introduced Superposed Decoding, which generates $k$ drafts with a single autoregressive inference pass and predicts $k^2$ drafts per step using n-gram interpolation to filter out incoherent outputs.

To address low predictive accuracy in draft models, especially with diverse text inputs and significant gaps between draft and target models, \cite{liu2023online} proposed Online Speculative Decoding. This method updates draft models based on user query data, improving predictions by adapting to query distributions. To minimize additional parameters or training for effective draft models, \cite{xia2024swift} introduced Swift, a plug-and-play speculative decoding solution that uses layer-skipping, leveraging the compact draft model by skipping intermediate layers of the target LLM.

Token tree verification combines multiple candidate draft sequences into a token tree, sharing prefixes, and applies a tree attention mask for efficient verification. \cite{miao2023specinfer} introduced SpecInfer, a tree-based parallel decoding algorithm that uses small speculative models to predict LLM outputs and organizes predictions into a token tree. This approach reduces latency and computational costs while maintaining model quality. \cite{cai2024medusa} proposed Medusa, an efficient method that enhances LLM inference by adding extra decoding heads to predict multiple tokens in parallel, reducing decoding steps through parallel processing. \cite{li2024eagle} introduced EAGLE, a speculative sampling framework that addresses feature-level autoregression uncertainty by incorporating a token sequence advanced by one time step, enabling efficient second-to-top-layer feature prediction. Subsequently, \cite{li2024eagle2} improved this with EAGLE-2, introducing a context-aware dynamic draft tree for more accurate draft modeling, leveraging well-calibrated draft models with closely approximated confidence scores.

To explore speculative decoding for LVLMs, \cite{gagrani2024speculative} enhanced inference efficiency in LVLMs, focusing on the LLaVA 7B model. \cite{jang2024lantern} address token selection ambiguity, where visual autoregressive models assign uniformly low probabilities to tokens, impairing speculative decoding. They propose LANTERN, a relaxed acceptance condition utilizing token interchangeability in latent space, restoring speculative decoding effectiveness while maintaining image quality and semantic coherence through a total variation distance bound. \cite{teng2024accelerating} introduced Speculative Jacobi Decoding (SJD) for autoregressive text-to-image generation, enabling the model to predict and accept multiple tokens per step, generating images more efficiently than traditional methods.
\section{Decoding Applications}
\label{sec:applications}
\begin{figure*}[!t]
    \centering
    \includegraphics[width=0.8\linewidth]{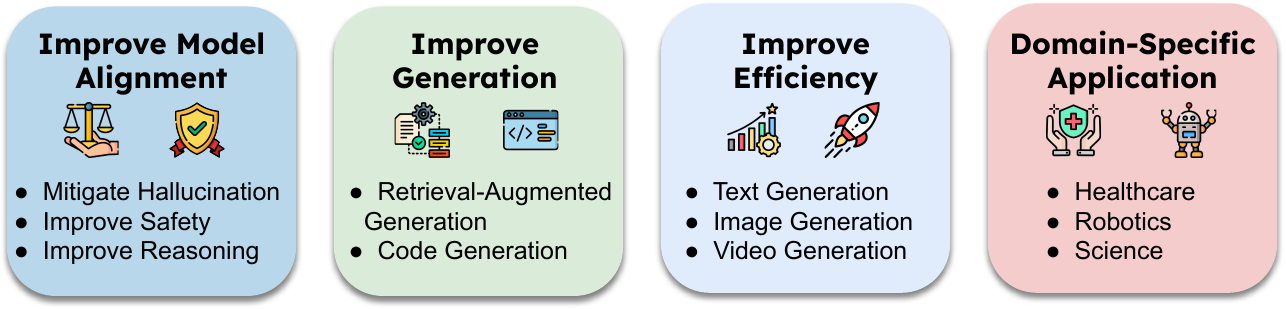}
    \caption{Decoding applications.}
    \label{fig:applications}
\end{figure*}

In this section, we shift our focus to organizing these methods according to the applications in which they have been used. Decoding algorithms play a crucial role across a variety of applications, from enhancing model alignment to optimizing specific generation tasks. Understanding how these algorithms are adapted to different applications provides valuable insights into their versatility and effectiveness. Figure~\ref{fig:applications} illustrates the diverse applications of decoding methods.

\subsection{Improve Model Alignment}
\label{sec:alignment}
Decoding strategies play a vital role in improving model alignment by mitigating hallucinations, enhancing safety, and strengthening reasoning capabilities. These strategies involve tailoring the output generation process during inference to achieve the desired outcomes. Unlike other methods of improving model alignment, decoding strategies provide a dynamic and adaptive approach, ensuring that the model consistently meets user expectations while adhering to ethical standards.

\subsubsection{Mitigate Hallucination}
Decoding methods have emerged as an effective inference-time tool for mitigating hallucinations \cite{huang2023survey}, which refers to the generation of plausible but factually incorrect content by the model. Compared to prompt-based \cite{gou2023critic, wang2023explainable, zhang2023knowledge} and knowledge-editing \cite{zhang2024truthx, li2024inference} approaches, decoding methods are model-agnostic and offer better interpretability. \cite{gema2024decore} introduced Decoding by Contrasting Retrieval Heads (DeCoRe), a decoding strategy that mitigates hallucinations by dynamically contrasting the outputs of a base and masked LLM, guided by conditional entropy. Similarly, \cite{yang2024improving} proposed a Comparator-driven Decoding-Time (CDT) framework, which generates hallucinatory and truthful comparators using multi-task fine-tuning and refines next-token predictions by contrasting logit differences between the target LLMs and these comparators.

In addition to mitigating hallucinations in LLMs, recent studies have demonstrated the effectiveness of contrastive decoding in addressing object hallucinations in LVLMs. \cite{lee2024delve} explores visual contrastive decoding techniques, such as image downsampling and editing, to reduce hallucinations. \cite{park2024convis} proposed ConVis, which reconstructs images from hallucinated captions using a text-to-image model and compares probability distributions to capture visual contrastive signals that penalize hallucination generation. To counter hallucinations caused by strong language model priors suppressing visual input, \cite{wang2024mllm} introduced DeCo, a dynamic correction decoding method that selectively integrates knowledge into the final layer to adjust output logits. \cite{huo2024self} developed CT$^2$S, which preserves only the least important vision tokens after early decoder layers, enhancing vision-text associations during autoregressive decoding. Inspired by the Information Bottleneck theory, \cite{kan2024catch} proposed CATCH, which integrates complementary visual decoupling for information separation, non-visual screening for hallucination detection, and adaptive token-level contrastive decoding for mitigation. More recently, \cite{chen2025attention} has discovered an ``attention hijacking'' phenomenon, where interference from instruction tokens distorts visual perception, diverting attention to less discriminative regions and leading to hallucinations. Additionally, \cite{tu2025attention} proposed an attention reallocation mechanism that redistributes excess attention from output tokens to visual tokens, reducing LVLMs’ reliance on language priors and strengthening visual input dependence to mitigate hallucinations.

\subsubsection{Improve Safety}
Besides mitigating hallucination, decoding methods are also used to enhance model safety. \cite{banerjee2024safeinfer} introduced Safeinfer, a context-adaptive safety alignment strategy for generating safe responses using safety-guided decoding, which selects tokens based on the safety-optimized distributions to ensure ethical content. Similarly, \cite{zeng2024root} proposed the Root Defense Strategy, a decoder-oriented defense that corrects harmful queries directly, rather than rejecting them outright. Recognizing that LLMs may appear to block harmful queries but still harbor latent risks, \cite{wang2024probing} proposed Jailbreak Value Decoding (JVD). JVD evaluates the probability of generating harmful content in subsequent decoding steps from any given point.

Instead of manually selecting contrastive models or instruction templates, \cite{zhao2024adversarial} proposed Adversarial Contrastive Decoding (ACD), an optimization-based framework that generates two opposing system prompts for prompt-based contrastive decoding. Using contrastive decoding, \cite{niu2024parameter} successfully mitigates toxicity in a parameter-efficient manner. To improve alignment through decoding, \cite{chakraborty2024transfer} uses Q-learning to adjust the response distribution directly, maximizing a target reward without requiring model updates. Moreover, \cite{bao2024decoding} identifies amplification bias and homogeneity issues in existing LLM decoding methods for recommendations. They propose Debiasing-Diversifying Decoding (D$^3$), which disables length normalization for ghost tokens to reduce amplification bias and uses a text-free assistant model to promote less frequent token generation, addressing recommendation homogeneity. To ensure the safety of conversational LLMs, \cite{furniturewala2024impact} examines the effect of various decoding methods on the alignment between LLM-generated and human conversations. They show that fewer beams in beam search and lower $P$ values in Nucleus sampling could lead to better alignment. PAD \cite{wang2025privacy} adaptively injects calibrated Gaussian noise into token logits to mitigate privacy leakage of retrieved context in RAG.

\subsubsection{Improve Reasoning}
Reasoning \cite{huang2022towards} is a key aspect of human intelligence and an important trait for LLMs. Recent works have shown that reasoning chains are embedded in token selection, and decoding methods can enhance reasoning ability. \cite{o2023contrastive} demonstrated that contrastive decoding improves reasoning in LLMs, outperforming existing methods by preventing abstract reasoning errors and avoiding simpler strategies like copying input sections during chain-of-thought reasoning. \cite{phan2024distillation} proposed Distillation Contrastive Decoding (DCD), which enhances LLM reasoning at inference time by employing Contrastive Chain-of-Thought Prompting and advanced distillation techniques, including Dropout and Quantization, without requiring expert and amateur models. By investigating top-k alternative tokens, \cite{wang2024chain} found that CoT paths often emerge within these sequences and can be elicited from LLMs by modifying the decoding process.

Recently, to improve the performance of models that have been knowledge-edited for reasoning questions, \cite{sun2024outdated} proposed Outdated Issue-Aware Decoding (DISCO), which captures the difference in the probability distribution between the original and edited models. To improve LLM reasoning in cross-domain settings, \cite{shen2024learning} proposed a method to teach multiple LLMs to collaborate by interleaving their generation at the token level. In parallel to syntactic decoding, such as auto-regressive decoding, \cite{peyrard2024era} introduced semantic decoding, a perspective that views collaborative processes as optimization procedures within a semantic space, offering an abstraction for search and optimization directly in the space of semantic tokens.

In addition to improving reasoning performance, decoding methods can enhance inference efficiency for reasoning tasks. To reduce the cost of generating a full CoT reasoning chain, \cite{liu2024expediting} introduced an auxiliary CoT model that generates and compresses the entire thought process into a compact token representation, semantically aligned with the original CoT output. To reduce the inference latency in tree-search-based reasoning methods, \cite{wang2024seed} introduced SEED, an efficient inference framework that accelerates reasoning tree construction using speculative scheduled execution, parallel drafting with speculative decoding, and a rounds-scheduled strategy to manage parallel drafts without verification conflicts. To reduce the cost of self-consistency decoding from the sampling process, \cite{chen2024self} proposed self-para-consistency, where multiple paraphrases are generated for each test question.

\subsection{Improve Generation Tasks}
\label{sec:specific}
Decoding strategies enhance large generative models in tasks such as retrieval-augmented generation and code generation, improving both output quality and efficiency. These advancements focus on refining decoding methods, optimizing model architectures, and incorporating external resources to enhance performance.

\subsubsection{Retrieval Augmented Generation}
Decoding methods can effectively improve RAG at inference time. \cite{qiu2024entropy} proposed entropy-based decoding to enhance truthfulness in retrieval-augmented LLMs, addressing challenges in ensuring faithful information retrieval. Similarly, adaptive contrastive decoding (ACD) \cite{kim2024adaptive} effectively incorporates contextual influence, improving the model's ability to generate contextually relevant outputs. To overcome the limitations of the generative retrieval model’s fixed parametric capacity, \cite{lee2022nonparametric} introduced Nonparametric Decoding (Np Decoding), which replaces standard embeddings with nonparametric contextualized vocab embeddings. Additionally, \cite{zeng2024planning} proposed PAG, an optimization and decoding approach that guides the autoregressive generation of document identifiers in generative retrieval models through simultaneous decoding, streamlining the document retrieval process. More recently, to speed up language model generation with a retrieval-based approach, \cite{he2023rest} proposed Retrieval-Based Speculative Decoding (REST). Unlike previous methods that rely on a draft language model, REST uses retrieval to generate draft tokens, leveraging the observation that text generation often follows common phrases and patterns.

\subsubsection{Code Generation}
Besides RAG, various decoding methods have been used for code generation with LLMs. \cite{zhu2023improving} introduced Adaptive Temperature (AdapT) sampling, which adjusts the temperature during token decoding: higher for challenging tokens to explore diverse options, and lower for confident tokens to reduce tail randomness noise. \cite{li2024doce} proposed Decoding Objectives for Code Execution (DOCE), a framework for execution-based evaluation. They focus on the effects of high-temperature sampling, execution-based reranking with high-quality unit tests, and self-debugging with multiple candidates. \cite{pimparkhede2024doccgen} introduced DocCGen, a two-step NL-to-code generation framework for structured domain-specific languages like YAML and JSON. It first detects relevant libraries using documentation to match the query, then constrains decoding with schema rules from these libraries.

To address security and correctness in code generation with Code LLMs, \cite{fu2024constrained} investigates a defense approach using constrained decoding to generate secure code, proving more effective than prefix tuning without requiring a specialized training dataset. \cite{wang2024uscd} presents Uncertainty-Aware Selective Contrastive Decoding (USCD), which improves one-pass code generation performance. It pre-judges noise in output distributions using standard deviation, then applies a ``lame'' prompt to reduce noise and enhance code quality. \cite{ni2023lever} proposed LEVER, which trains verifiers to assess program correctness using natural language input, the program, and its execution results. Programs are reranked by combining verification scores with LLM probabilities, marginalizing over identical execution results.

Additionally, decoding methods can help mitigate hallucination during code generation. \cite{nie2024decoding} proposed DESEC, a two-stage method that uses token-level features to guide decoding. It builds an offline token scoring model with a proxy Code LLM and adjusts token likelihoods during decoding based on these scores. \cite{tian2024selective} introduced Selective Prompt Anchoring (SPA), which addresses self-attention dilution in LLMs for code generation. SPA strengthens the influence of selected parts of the initial prompt by adjusting the logit distribution based on the difference between anchored and non-anchored text.

\subsection{Improve Generation Efficiency}
\label{sec:efficiency}
Decoding methods are used to boost generation efficiency across multiple domains, such as text, image, and video.

\subsubsection{Text Generation}
Beyond speculative decoding, several works focus on enhancing text generation efficiency through innovative decoding strategies. Reward-Augmented Decoding (RAD) \cite{deng2023reward} used a lightweight unidirectional reward model to guide a language model in producing text with desired properties. \cite{zhu2024hierarchical} introduced Hierarchical Skip Decoding (HSD), an efficient autoregressive method that adaptively skips decoding layers based on sequence length, reducing computational overhead without requiring additional trainable components. \cite{yang2023frustratingly} proposed Frustratingly Simple Decoding (FSD), which uses an anti-language model to penalize repetitive content. The anti-LM can be implemented as an n-gram model or a vectorized variant, adding no extra parameters and incurring minimal cost, making it as fast as greedy search.

More recently, \cite{liu2024adaptive} proposed ADED, a decoding methodology that enhances LLM efficiency without fine-tuning, utilizing an adaptive draft-verification process. \cite{su2025seesaw} proposed Seesaw, which uses dynamic model resharding to enable reconfiguration of parallelization strategies during decoding. \cite{fan2025position} introduced Position-Aware Depth Decay Decoding, which employs a power-law decay function to optimize the number of layers retained during token generation for more efficient performance.

\subsubsection{Image Generation}
To accelerate autoregressive text-to-image generation, \cite{teng2024accelerating} introduced Speculative Jacobi Decoding (SJD), a training-free probabilistic parallel decoding algorithm. \cite{feng2023emage} explores non-autoregressive text-to-image models that generate image tokens in parallel. They introduce an iterative mask-predict approach, enabling the model to refine its predictions using partially observed tokens, which enhances convergence speed and output quality. \cite{tang2024hart} introduced the Hybrid Autoregressive Transformer (HART), an autoregressive visual generation model that employs hybrid tokenization. This approach enables continuous feature decoding during generation, overcoming the limitations of finite VQ codebooks and improving overall generation quality.

\subsubsection{Video Generation}
By reformulating dense caption generation as a set of prediction tasks, \cite{wang2021end} proposed PDVC, an end-to-end dense video captioning framework with parallel decoding. To ensure global coherence and local realism in video generation, GLOBER \cite{sun2024glober} uses a video decoder that processes global features and synthesizes video frames in a non-autoregressive manner. For enhanced flexibility, the video decoder incorporates normalized frame indexes to perceive temporal information, enabling the generation of arbitrary sub-video clips with predefined starting and ending frame indexes. Similarly, VideoGen \cite{li2023videogen} employs an advanced video decoder trained on unlabeled data to produce high-definition videos with strong temporal consistency and high frame fidelity.

\subsection{Domain-Specific Applications}
\label{sec:domain}
Decoding strategies can significantly enhance domain-specific applications, such as healthcare and robotics. In the \textit{healthcare} domain, \cite{xu2024mitigating} addresses hallucination in medical information extraction tasks with Alternate Contrastive Decoding (ALCD). They redefine the task as an identification-and-classification process, separating these steps by masking token optimization during fine-tuning. During inference, ALCD improves both identification and classification by contrasting output distributions from sub-task models, thereby minimizing interference from other LLM capabilities. Additionally, an adaptive constraint strategy refines the contrastive token scope, further enhancing performance. In the scientific domain, \cite{zheng2024llm4brain} implements cross-subject semantic decoding for video-stimulated fMRI, while \cite{chen2024open} explores open-vocabulary auditory neural decoding using fMRI-prompted LLMs.

In \textit{robotics}, \cite{huang2024grounded} formulates the construction of action sequences as a probabilistic filtering problem, ensuring that the sequences are both probable according to the LM and feasible within grounded environmental models. Their approach demonstrates how grounded models can be derived from both simulation and real-world domains. By integrating knowledge from language models and grounded models, their decoding strategy effectively addresses complex, long-horizon embodiment tasks, enabling the generation of accurate and feasible action sequences. Similarly, \cite{liu2024bidirectional} introduced Bidirectional Decoding (BID), an inference algorithm that combines action chunking with closed-loop operations. BID samples multiple predictions at each time step, optimizing for backward coherence (alignment with prior decisions) and forward contrast (high likelihood of future plans). This dual optimization approach ensures consistent decision-making while allowing the system to adapt to unexpected environmental changes.
\section{Discussions}
\label{sec:challenges}
Despite the significant potential of decoding methods across various tasks and applications, several challenges remain. 
In this section, we highlight the limitations of current decoding methods and discuss possible future research directions.

\subsection{Exploring Dynamic and Universal Decoding Methods}
Although decoding methods are effective for specific tasks, they often rely on manually crafted examples. As discussed earlier, token-wise and layer-wise contrastive decoding (\S~\ref{sec:contrastive}) enhance control over text generation in LLMs and LVLMs. However, their effectiveness heavily depends on the selection of contrastive examples or layers. For instance, \cite{li2022contrastive} compares outputs from smaller language models to those from larger ones, assuming that bigger models produce higher-quality text. However, this assumption does not always hold, as there are cases where the generation is worse with CD. Moreover, \cite{sharma2023towards} even points out that smaller LLMs tend to be less sycophantic, meaning they are less likely to prioritize aligning with user beliefs over providing truthful responses. This highlights the complexity of crafting contrastive examples, as it requires balancing multiple qualities in the generated text. Similarly, in selecting contrasting layers, the optimal layers in DoLa \cite{chuang2023dola} are sensitive across datasets, making it less versatile since it requires a task-specific validation set. This limitation presents an opportunity for future research to develop methods for constructing dynamic, universally applicable contrastive examples.

For guided decoding (\S~\ref{sec:guided}), search methods like the one used in \cite{zhang2023planning}, rely on test cases and are constrained by a small search space. While \cite{feng2023alphazero} demonstrated that TS-LLMs perform well across reasoning, planning, alignment, and decision-making tasks on trees with a depth of 64, they still struggle to scale to larger scenarios due to the computational overhead introduced by node expansion and value evaluation. This highlights the need for more efficient strategies that can scale to larger problem spaces without introducing excessive computational costs.

\subsection{Interpreting Decoding Methods}
While some decoding methods, such as those explored by \cite{wang2024chain}, offer insights into the intrinsic reasoning abilities of LLMs through the lens of decoding, more theoretical foundations are needed to understand why models behave in certain ways during the decoding process. For instance, \cite{shi2024thorough} underscores the critical role of hyperparameter tuning in optimizing decoding methods. Their findings highlight that while some approaches can achieve impressive performance, they often require substantial effort to dial in the hyperparameters. In an initial effort, \cite{chang2024explaining} theoretically demonstrates that contrastive decoding can be viewed as linearly extrapolating the next-token logits from a large, hypothetical language model. They find that this linear extrapolation may prevent contrastive decoding from producing the most obvious answers, as these are already assigned high probabilities by the base LM. 

Moving forward, future research should focus on exploring the interpretability of these methods, which could provide valuable insights into how different decoding strategies align with human reasoning and decision-making processes. One potential approach is \textit{mechanistic interpretability} \cite{bereska2024mechanistic, lin2025survey, belrose2023eliciting}, which seeks interpretability by reverse-engineering black-box models. For instance, \cite{pal2023future} investigates the possibility of decoding multiple future tokens from a single token's hidden representation. Their causal intervention study reveals that certain layers can approximate the model's output with up to 48\% accuracy from a single hidden state, providing significant insights into the model’s prediction chain at each hidden state.

\subsection{Combining Different Decoding Paradigms}
As discussed in Section~\ref{sec:paradigms}, contrastive decoding, guided decoding, and parallel decoding are versatile paradigms that have proven effective for a variety of tasks, such as controlling generation, improving quality, and enhancing efficiency. Given their individual successes, it is natural to explore the potential of combining these paradigms. For instance, \cite{yuan2023speculative} combined Speculative Decoding with contrastive decoding, achieving both improved generation quality and inference speedup. Additionally, these decoding paradigms can be integrated with other advanced generation techniques, such as RAG. \cite{wang2024speculative} enhanced RAG by incorporating drafting with a set of specialized drafters, which provide diverse perspectives on the evidence while reducing the token count per draft. Future research could investigate further combinations of decoding paradigms to address the unique challenges posed by LLMs.

\subsection{Expanding the Diversity of Decoding Objectives}
Traditionally, decoding methods for LMs have primarily focused on text generation quality. However, recent works on LLMs and LVLMs have shifted toward improving the alignment of generated outputs. While several aspects of alignment have been explored, including toxicity \cite{liu2021dexperts}, truthfulness \cite{shi2023trusting, chuang2023dola}, and safety \cite{xu2024safedecoding, zhong2024rose}, significant gaps remain in addressing other critical issues, such as \textit{privacy, bias, and copyright concerns}. Additionally, while decoding methods have been applied across various domains, there are limited applications in high-stakes sectors like healthcare, law, and finance. Finally, for works focusing on improving the decoding efficiency through parallel decoding, there is a need to ensure the balance of decoding accuracy and efficiency. As noted by \cite{xia-etal-2024-unlocking}, there is still room for improvement to align the drafter with the target LLM for speculative decoding to scale up the drafter to improve decoding accuracy while maintaining its efficiency advantages. Future research could explore leveraging decoding methods to tackle these issues, offering a plug-and-play solution to enhance the trustworthiness of LLMs for important applications.

\subsection{Improving Adversarial Robustness}
Decoding methods are often overlooked from the security perspective. While much of the research focuses on enhancing the security of LLMs through techniques such as preventing data leakage and defending against jailbreaking attacks, the decoding mechanism itself is often neglected as a potential security risk. Decoding methods like SafeDecoding \cite{xu2024safedecoding} and ROSE \cite{zhong2024rose} are designed to generate safe responses from models, but they can also be exploited by attackers to generate malicious outputs. More concerningly, \cite{naseh2023stealing} demonstrated that adversaries with typical API access can steal the type and hyperparameters of a model's decoding algorithm at a low cost. This highlights a growing security concern regarding potential attacks that manipulate the underlying decoding method of a model. Moving forward, it would be valuable to explore strategies for defending against decoding-based attacks. As suggested by \cite{naseh2023stealing}, watermarking \cite{szyller2021dawn} could serve as a potential countermeasure, adding noise to the final probability distribution and making it more difficult for attackers to extract hyperparameters from the target model.

\section{Conclusion}
This survey provides a comprehensive review of three primary decoding paradigms and their diverse applications in LLMs and LVLMs, showcasing their effectiveness and efficiency in tackling complex generation tasks. 
We believe that decoding methods offer a cost-effective way to enhance and extend LLM capabilities and hope our work sparks further discussion and research in this area. 

\section*{Acknowledgments}
This material is based upon work supported by NSF awards (SaTC-2241068, IIS-2506643, and POSE-2346158), a Cisco Research Award, and NSF NAIRR Pilot Award \#240469. The views and conclusions contained in this document are those of the authors and should not be interpreted as necessarily representing the official policies, either expressed or implied, of the National Science Foundation. This work is supported in part by NSF under grants III-2106758.

%
\bibliographystyle{abbrv}
\bibliography{anthology, custom}  
%
%

\end{document}